\pdfoutput=1
\documentclass[sigconf,nonacm]{acmart}
\usepackage{subcaption}
\usepackage{graphicx}
\usepackage{enumitem}
\usepackage{multirow}
\usepackage{soul}
\usepackage{stfloats}
\usepackage{makecell}
\usepackage[ruled,boxed,linesnumbered]{algorithm2e}
\usepackage{tabularx}
\newcounter{problemname}
\newcounter{definitionname}

\newcommand{\eat}[1]{}

\AtBeginDocument{
  }

\graphicspath{ {./} }

\begin{document}

\title{TempoGPT: Enhancing Time Series Reasoning\\via Quantizing Embedding}

\author{Haochuan Zhang}
\affiliation{%
		\department{School of Automation}
	\institution{Central South University}
	\country{Changsha, China}
}
\email{244607004@csu.edu.cn}

\author{Chunhua Yang}
\affiliation{%
		\department{School of Automation}
	\institution{Central South University}
	\country{Changsha, China}
}
\email{ychh@csu.edu.cn}

\author{Jie Han}
\affiliation{%
		\department{School of Automation}
	\institution{Central South University}
	\country{Changsha, China}
}
\email{hanjie@csu.edu.cn}
\authornote{Jie Han is the Corresponding Author.}

\author{Liyang Qin}
\affiliation{%
		\department{School of Automation}
	\institution{Central South University}
	\country{Changsha, China}
}
\email{qly520@csu.edu.cn}

\author{Xiaoli Wang}
\affiliation{%
		\department{School of Automation}
	\institution{Central South University}
	\country{Changsha, China}
}
\email{xlwang@csu.edu.cn}

\begin{abstract}
Multi-modal language model has made advanced progress in vision and audio, but still faces significant challenges in dealing with complex reasoning tasks in the time series domain. The reasons are twofold. First, labels for multi-modal time series data are coarse and devoid of analysis or reasoning processes. Training with these data cannot improve the model’s reasoning capabilities. Second, due to the lack of precise tokenization in processing time series, the representation patterns for temporal and textual information are inconsistent, which hampers the effectiveness of multi-modal alignment.
To address these challenges, we propose a multi-modal time series data construction approach and a multi-modal time series language model (TLM), TempoGPT. Specially, we construct multi-modal data for complex reasoning tasks by analyzing the variable-system relationships within a white-box system. Additionally, proposed TempoGPT achieves consistent representation between temporal and  textual information by quantizing temporal embeddings, where temporal embeddings are quantized into a series of discrete tokens using a predefined codebook; subsequently, a shared embedding layer processes both temporal and textual tokens.
Extensive experiments demonstrate that TempoGPT accurately perceives temporal information, logically infers conclusions, and achieves state-of-the-art in the constructed complex time series reasoning tasks.  Moreover, we quantitatively demonstrate the effectiveness of quantizing temporal embeddings in enhancing multi-modal alignment and the reasoning capabilities of TLMs. Code and data are available at \href{https://github.com/zhanghaochuan20/TempoGPT}{https://github.com/zhanghaochuan20/TempoGPT}.
\end{abstract}


\maketitle
\section{Introduction}
Multi-modal language models can leverage the robust semantic understanding and reasoning capabilities of large language models (LLMs) to facilitate the processing of diverse modalities and support multiple downstream tasks \cite{zhu2023minigpt, ansari2024chronos, deshmukh2023pengi, alayrac2022flamingo, liu2024visual}. Time series data, a ubiquitous form of modal information, is essential in numerous domains, such as anomaly detection \cite{blazquez2021review}, medical diagnosis \cite{dudukcu2023temporal}, and financial economics \cite{liu2023financial}. Consequently, numerous researchers aim to take use of LLMs to improve the effectiveness of time series analysis \cite{minaee2024large,liu2023itransformer, zeng2023transformers}.

\begin{figure}[h]
	\centering
	\begin{subfigure}[t]{0.44\linewidth}
		\includegraphics[width=\linewidth]{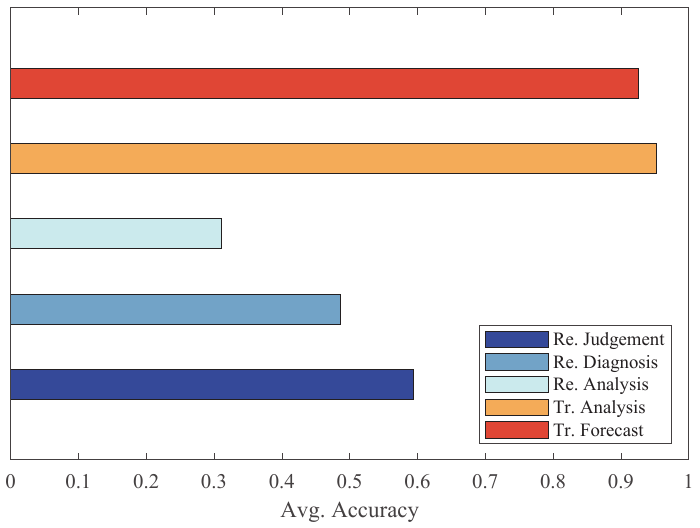}
		\caption{}
	\end{subfigure}
	\hfill
	\begin{subfigure}[t]{0.48\linewidth}
		\includegraphics[width=\linewidth]{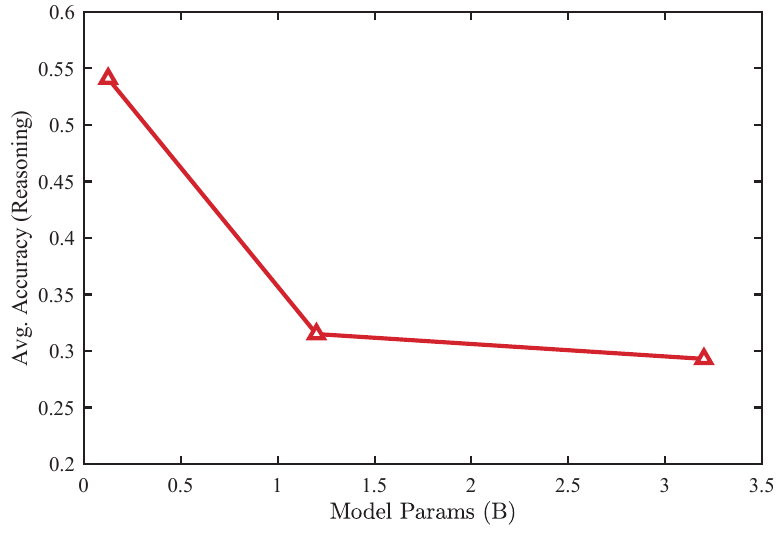}
		\caption{}
	\end{subfigure}
	\caption{(a) TLMs perform well in tasks related to trend (Tr.) but poorly in tasks involving complex reasoning (Re.). (b) as the model size increases, the performance of TLMs deteriorates on reasoning tasks.}
	\label{fig:intro_trend_reasoning}
\end{figure}

However, as illustrated in Figure \ref{fig:intro_trend_reasoning}, when introducing the reasoning capabilities of LLMs to the time series domain, while multi-modal time series language models (TLMs) perform well on trend-related tasks, they underperform in complex reasoning tasks and struggle to effectively utilize LLMs' reasoning capabilities \cite{merrill2024language}.  Despite extensive research on TLMs and numerous efforts to leverage advanced model frameworks and training methods, the performance of TLMs in time series reasoning has yet to show significantly improvement \cite{chow2024timeseriesreasoningllms}. This indicates that it is necessary to identify and address the fundamental issues underlying the limitations of TLMs in complex reasoning tasks.

\begin{figure}[ht]
	\centering
	\includegraphics[width=1\linewidth]{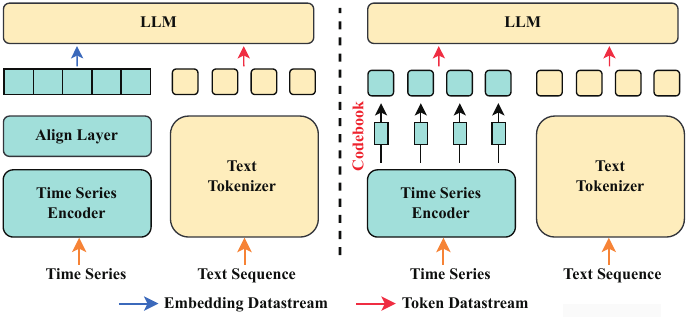}
	\caption{On the left side of the diagram, due to the lack of temporal tokenizers, TLMs process inputs with inconsistent representation patterns. On the right side of the diagram,  proposed TempoGPT quantizes temporal embeddings with a predefined codebook, ensuring a precise tokenization and consistent representation pattern.}
	\label{fig:compare-encoding}
\end{figure}

By analyzing the crucial technologies in constructing TLMs, there are two significant challenges.
\textit{\textbf{1) Multi-modal data.}}  The community lacks high-quality multi-modal time series data as the foundation for training TLMs \cite{zhou2023one, cai2024timeseriesexam, wang2024chattimeunifiedmultimodaltime, xie2024chatts}. Most datasets use basic classification labels, such as diagnoses like "sinus rhythm" and "sinus irregularity" in electrocardiogram (ECG) data \cite{li2024frozen}. Those datasets lack a thorough analysis of complex relationships, such as those between variables and between variables and physical systems \cite{li2024urbangpt}. Training with those data turns TLMs into classifiers, rather than improving their semantic understanding and reasoning capabilities. 
\textit{\textbf{2) Multi-modal alignment.}} As shown in the left side of Figure \ref{fig:compare-encoding}, unlike text sequences where individual words can be tokenized into discrete tokens \cite{mikolov2013efficient}, temporal representation methods lack precise tokenization techniques due to the absence of inherent semantic information in time series data points and the infinite variability of patches \cite{nie2022time}. Those limitations result in inconsistent representation patterns for textual and temporal information in TLMs. As Figure \ref{fig:intro_trend_reasoning} illustrates, while this issue has minimal impact on basic time series analysis tasks for TLMs, it poses challenges when strong multi-modal alignment and reasoning capabilities are required.

To effectively address the challenges faced by TLMs in complex reasoning tasks, it is crucial to tackle the aforementioned issues. Considering the significance of multi-modal data involving complex analysis and reasoning processes in improving the reasoning capabilities of TLMs, we advocate for generating time series data within a white-box physical system. In this system, the relationships between variables and the system can be systematically analyzed  using rule-based methods and human-in-the-loop strategies, thereby facilitating the creation of comprehensive multi-modal data. Furthermore,  considering the potential risks posed by inconsistent information representation patterns to the reasoning capabilities of TLMs, as shown in the right side of Figure \ref{fig:compare-encoding}, we advocate for quantizing temporal embeddings into discrete tokens.  This enables a precise tokenization and facilitates the capture of inherent semantic information of time series \cite{chung2023text, cheng2024advancingtimeseriesclassification}. These discrete tokens can be integrated into the vocabulary of LLMs and subsequently processed alongside textual tokens by a shared embedding layer, thus achieving a consistent representation pattern with textual information. In this way, temporal information can be better aligned with textual information, thereby enhancing the reasoning capabilities of TLMs.

Based on the above motivations, this paper presents a multi-modal time series data construction method and a multi-modal time series language model, TempoGPT.  Our contributions are primarily reflected in the following aspects:

\begin{itemize}[leftmargin=*, topsep=1pt]
	\item \textit{\textbf{Multi-Modal Time Series Data.}}  We propose a novel multi-modal data construct approach, which enables effective analysis of variable-system relationships and facilitates the creation of multi-modal data for time series reasoning tasks.
	\item \textit{\textbf{Time Series Language Model.}} We design a multi-modal time series language model, TempoGPT. TempoGPT quantizes temporal embeddings into discrete tokens and expands the embedding layer to process both temporal and textual tokens, achieving a consistent representation of temporal and textual information.
	\item \textit{\textbf{Time Series Reasoning.}}  Experimentally, TempoGPT achieves state-of-the-art on constructed time series reasoning tasks. Additionally, we extensively analyze the factors that influence the time series reasoning capabilities of TLMs and demonstrate the effectiveness of quantizing temporal embeddings in enhancing multi-modal alignment and reasoning capabilities.
\end{itemize}

\section{RELATED WORK}
\subsection{Time Series Language Model}
The remarkable capabilities of LLMs attract numerous researchers to explore how LLMs' reasoning capabilities can enhance the effectiveness of time series analysis  \cite{zhang2024largelanguagemodelstime, gruver2023large, rasul2023lag,caotempo}.

Initially, LLMs are utilized as innovative tools for specific time series analysis tasks, such as forecasting and classification, by designing specialized encoders and output heads \cite{suntest,  jin2023time, liu2024autotimes}. However, these studies do not effectively utilize the natural language processing and reasoning capabilities of LLMs \cite{tan2024languagemodelsactuallyuseful, zheng2024revisited}.
Subsequently, some studies attempt to utilize the reasoning capabilities of LLMs through alignment  or prompt-based methods, without altering the output heads of the LLMs \cite{cai2023jolt, tang2024time, mirchandani2023large}. Although prompt-based methods do not add extra components to LLMs, they struggle to scale to multivariate time series analysis due to limit input tokens. Furthermore, challenged with coarse multi-modal data labels, both alignment and prompt-based methods are limited to fundamental time series analysis tasks. Recent studies successfully utilize TLMs to analyze fundamental characteristics of time series \cite{wang2024chattimeunifiedmultimodaltime, xie2024chatts}, such as trend and seasonality. However, research on TLMs in complex time series reasoning tasks is insufficient, and their performance is currently suboptimal \cite{merrill2024language}.

\subsection{Multi-modal Alignment}
Multi-modal alignment methods, such as BLIP-2 \cite{li2023blip} and CLIP \cite{radford2021learning}, are widely studied in the visual domain, and demonstrate outstanding performance \cite{li2022blip, liu2024improvedbaselinesvisualinstruction, girdhar2023imagebind}.  
By emulating alignment methods from the visual domain, researchers achieve notable effectiveness in specific time series analysis tasks. However, they struggle to scale these methods to  complex reasoning tasks, which place high demands on multi-modal alignment and the reasoning capabilities of TLMs \cite{merrill2024language,chow2024towards}.
Observing inconsistencies in the representation patterns between temporal and textual information, some studies utilize quantization-based methods to tokenize temporal embeddings and unify the representation pattern \cite{duan2023dewave, chung2023text}. These studies primarily concentrate on specific tasks in time series analysis, where the extraction of temporal information is more critical than multi-modal alignment, resulting in insignificant benefits from a consistent representation pattern \cite{haoyietal-informer-2021}. 
However, for the time series reasoning tasks that require strong  alignment between temporal and textual information, current studies seldom explore whether inconsistent representation patterns hinder TLMs' reasoning capabilities.

\section{Multi-modal Data Generation} \label{Multi-modal Data Generation}
\begin{figure*}[ht]
	\centering
	\includegraphics[width=1\linewidth]{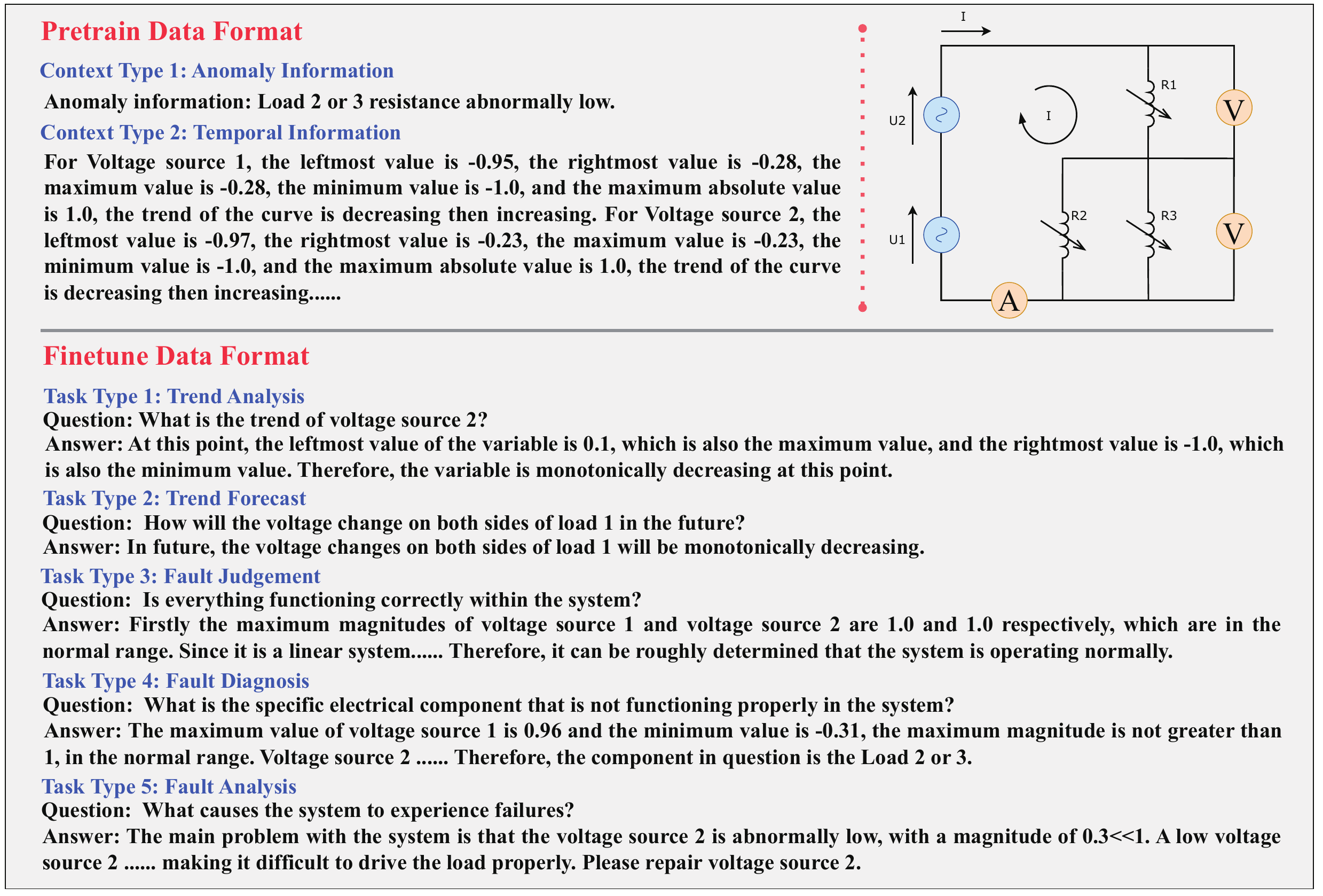}
	\caption{At the top of the diagram, the left side presents the pre-training data components, which include anomaly and temporal information, while the right side presents the electrical simulation model responsible for generating the multi-modal data. On the bottom of the diagram, the construction of fine-tuned data is showcased, primarily comprising two trend-related tasks: trend analysis and trend forecasting; along with three reasoning-related tasks: fault judgment, fault diagnosis, and fault analysis. Due to space limitations, we present detailed data in the Appendix \ref{apx: data}. }
	\label{fig:pretrain_data}
\end{figure*}
Considering the scarcity of multi-modal time series data and the insufficient attention to the complex analysis and reasoning process in time series,  we suggest simulating multi-modal data within a white-box system with known mechanisms, generating labels through rule-based and human-in-the-loop approaches, and using ChatGPT \cite{wu2023brief} to enhance the diversity of the data further ultimately. In this manner, the relationships between time series and physical systems can be effectively explored and analyzed.

Initially, as shown on the right side of Figure \ref{fig:pretrain_data}, a basic linear circuit is constructed. The circuit includes two AC voltage sources with the same maximum  amplitude but different phases. There are three loads within the circuit: Load 1 is connected in series in the main loop, while Load 2 and Load 3 are connected in parallel and then in series with Load 1. The voltages across Voltage Source 1 and 2, Load 1, 2, and 3, as well as the current in the main loop, are observable. Additionally, to simulate abnormal operating conditions of the circuit, intrinsic parameters of the components within the circuit are randomly adjusted within a specific range, such as modifying the maximum amplitude of the voltage sources and the resistance of the loads. Time series data, consisting of six electrical variables, are generated based on the constructed electrical system: the voltage across Voltage Source 1, the voltage across Voltage Source 2, the total electromotive force of the circuit (sum of the voltages of Voltage Source 1 and 2), the voltage across Load 1, the voltage across Loads 2 and  3, and the current in the main loop of the circuit. 

Subsequently, we develop a series of templates for textual data to be used in both pre-training and fine-tuning stages. Textual data are derived from these templates by analyzing the generated time series data with rule-based and human-in-the-loop methods. The pre-training data comprise two main components: anomaly information related to electronic components and temporal information concerning the aforementioned six electrical variables. In the fine-tuning stage, we implement a chain-of-thought (CoT) \cite{wei2022chain} approach to simulate the human reasoning process and develop five types of time series reasoning tasks: trend analysis, trend forecast, fault judgment, fault diagnosis, and fault analysis. Trend analysis and trend forecast primarily involve the analysis of time series data, while the other three tasks concentrate on exploring the connections between time series and physical systems, as well as the complex reasoning involved.
Finally, to enhance the capability of TLMs in instruction following, ChatGPT is tasked with generating more detailed and complex queries based on the `Question' component of the fine-tuning data. 

We use the data constructed by the aforementioned method to train the proposed TempoGPT, and test it on five types of time series reasoning tasks to evaluate TempoGPT's reasoning capabilities.

\section{TempoGPT}
\subsection{Architecture}
\begin{figure*}[ht]
	\centering
	\includegraphics[width=1\linewidth]{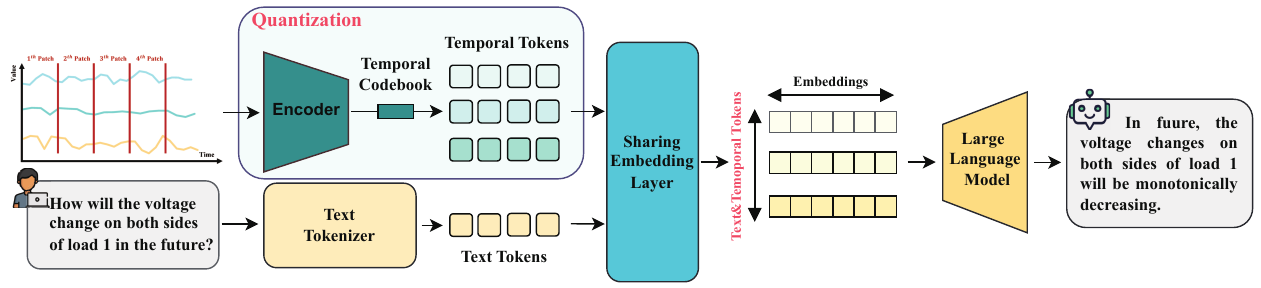}
	\caption{TempoGPT network architecture. TempoGPT employs quantization encoding to tokenize time series into temporal tokens. A shared embedding layer processes these tokens alongside textual tokens to  achieve consistent representation pattern and generate corresponding embeddings. These embeddings are then processed by LLMs to generate the final text response.}
	\label{fig:pattern}
\end{figure*}

TempoGPT is constructed to achieve a consistent representation pattern between temporal and textual information,  and to accomplish complex time series reasoning tasks.  As shown in Figure \ref{fig:pattern}, TempoGPT involves three key technologies: 1. quantization encoding; 2. shared embedding layer. 3. large language model.
\subsubsection{\textbf{Quantization Encoding}}
The purpose of quantization encoding is to tokenize time series into discrete tokens utilizing a predefined encoder and temporal codebook.  The encoder, which is based on a weight-sharing 1-D convolutional network, transforms the time series into high-dimensional temporal embeddings.  Furthermore, we employ patching and channel-independence techniques to enable the encoder to better represent the temporal information \cite{nie2022time}. The codebook consists of a fixed-size set of codewords, where each codeword is represented as a vector in the discrete latent space and is associated with a corresponding token (index) \cite{oord2018neuraldiscreterepresentationlearning}. Each of the aforementioned temporal embeddings is then mapped to the nearest discrete codeword based on the predefined codebook, thereby forming a temporal token.  This would provide a precise tokenizer for the processing of time series.
Notably, the encoder and  temporal codebook in the quantization component can be developed by training a Vector Quantised{-}Variational AutoEncoder (VQ-VAE), as detailed in \cite{cheng2024advancingtimeseriesclassification}. In other words, we load the encoder and codebook of VQ-VAE in the quantization component and freeze these parameters during the subsequent training process of TempoGPT. 

Specifically, for a  time series $X = (\mathbf{x_1},\mathbf{ x_2}, ... , \mathbf{ x_M}) \in \mathbb{R}^{M \times N}$ with $M$ timestamps and $N$ variables, we firstly encode it into $D$-dimensional temporal embeddings  $X_E = (\mathbf{x_E^1}, \mathbf{x_E^2}, ... , \mathbf{x_E^P}) \in \mathbb{R}^{P \times N \times D}$, where $P$ is the number of patches of each variable. Subsequently, each embedding vector in $X_E$ is converted into a temporal token using the predefined codebook, forming $X_T \in \{\text{<0>,<1>,<2>, ..., <K>}\}$. Through this process, quantization encoding effectively transforms the input from each patch of every variable into a temporal token.

\subsubsection{\textbf{Shared Embedding Layer}}
The shared embedding layer functions to extend the LLMs' embedding layer to process both textual and temporal tokens simultaneously, thereby ensuring a consistent representation pattern for textual and temporal information. Temporal tokens are added to the existing vocabulary of LLMs, and the word embedding matrix is correspondingly expanded to ensure compatibility with the vocabulary.  

Assuming the length of the original vocabulary $V_0$ of the LLM is $|V_0|$, the vocabulary $V_1$, consisting of additional temporal tokens and special tokens, has a length of $|V_1|$, and $V_0 \cap V_1 = \varnothing$. The original vocabulary $V_0$ is expanded by merging it with $V_1$ to ultimately form  $V_2$.

\begin{equation}
	\begin{split}
		V_2 &= V_0 \cup V_1.
	\end{split}
\end{equation}

While expanding the vocabulary, it is necessary to simultaneously expand the corresponding word embedding matrix to ensure that the dimensions of the vocabulary and the embedding matrix are compatible.
Assume that the word embedding matrices corresponding to $V_0$, $V_1$ and $V_2$ are $W_0 \in \mathbb{R}^{|V_0| \times d}$, $W_1 \in \mathbb{R}^{|V_1| \times d}$ and $W_2 \in \mathbb{R}^{|V_2| \times d}$, respectively, where $d$ is the dimensions of word embedding. According to Eq.\ref{eq:w2}, expand $W_0$  to derive $W_2$.
\begin{equation}
	\label{eq:w2}
	\begin{split}
		W_2 &= [W0;W1].
	\end{split}
\end{equation}

Subsequently, the process of representing information is the same for both text and time series data:  initial textual/temporal data is converted into textual/temporal tokens, each of which is mapped to a specific textual/temporal embedding vector through the shared embedding layer.
In this way, TempoGPT can represent temporal information by combining a finite set of discrete embedding vectors. Adjusting these finite embedding vectors allows for better alignment of temporal and textual information,  which is much more straightforward than representing temporal information in the  continuous embedding space.

\subsubsection{\textbf{Large Language Models}}
Finally, the constructed textual and temporal embedding vectors are fed into the main body of LLMs. LLMs further integrate and process temporal and textual information, ultimately generating the corresponding text response.

TempoGPT aims to explore whether a consistent pattern for temporal and textual representation can enhance performance on time series reasoning tasks, positioning it as a general model framework. Consequently, we study  five types of base models in TempoGPT, including GPT-2 (125M) \cite{radford2019language}, LLaMA-3.2-1B (1.2B) \cite{Llama-3.2-1B}, LLaMA-3.2-3B (3.2B) \cite{Llama-3.2-3B}, tiny-llama (1.1B) \cite{zhang2024tinyllamaopensourcesmalllanguage}, and Phi-2 (2.7B) \cite{javaheripi2023phi}.

\subsection{Training} \label{training}
We train TempoGPT on our constructed electrical multi-modal time series dataset through pre-training and fine-tuning stages. The primary goal of TempoGPT during pre-training stage is to align temporal with textual information, while its fine-tuning stage focuses on enabling it to follow user instructions and process complex time series reasoning tasks. We do not discuss the training details of VQ-VAE here, assuming it has already been trained on time series data. For more details, please refer to \cite{cheng2024advancingtimeseriesclassification}. 

\subsubsection{\textbf{Pre-training}}
During the pre-training stage, only the parameters $W$ of the embedding layer in TempoGPT are adjusted, while the parameters of the remaining parts are frozen. This approach helps preserve LLMs' inherent capabilities while aligning textual information with temporal information. A sample in the pre-training dataset  consists of the pair ($X$, $\mathbf{y}$), where $X$ is an input time series with $M$ timestamps and $N$ variables, and $\mathbf{y}$ is the corresponding text labels,  as shown in Figure \ref{fig:pretrain_data}. The probability of the target response $\mathbf{y}$ during the pre-training stage can be calculated as follows:
\begin{equation}
	\begin{split}
		P(\mathbf{y}|X) = \prod_{i=1}^{|\mathbf{y}|} p_W(\mathbf{y}_i|\mathbf{y}_{<i};X).
	\end{split}
\label{eq:p_pre}
\end{equation}
where $\mathbf{y}_i$ indicates the $i$-th token in $\mathbf{y}$; $|\mathbf{y}|$ is the number of tokens in $\mathbf{y}$; $\mathbf{y}_{<i}$ indicates the tokens before $\mathbf{y}_i$ in $\mathbf{y}$.

\subsubsection{\textbf{Fine-tuning}}
During the fine-tuning stage, the embedding layer parameters $W$ and the main body of LLMs' parameters $\varepsilon$ in TempoGPT are adjusted, which can help TempoGPT better adapt to multiple downstream tasks.  Notably, when the base model is GPT-2, we employ full fine-tuning; otherwise, we utilize LoRA (Low-Rank Adaptation) for fine-tuning \cite{hu2021loralowrankadaptationlarge}. Additionally, to enhance the instruction-following capabilities of TempoGPT during the fine-tuning stage, the data sample is constructed as (($X$,$\mathbf{q}$),$\mathbf{y}$), where $\mathbf{q}$ represents the user command, randomly selected from the five types of time series reasoning tasks. The probability of the target response $\mathbf{y}$ during the fine-tuning stage can be calculated according to Eq.\ref{eq:fine-tuning}
\begin{equation}
	\begin{split}
		P(\mathbf{y}|X;\mathbf{q}) = \prod_{i=1}^{|\mathbf{y}|} p_{(W,\varepsilon)}(\mathbf{y}_i|\mathbf{y}_{<i};X;\mathbf{q}).
	\end{split}
\label{eq:fine-tuning}
\end{equation}

\section{Experiments}
In this section, we evaluate the time series reasoning capabilities of TempoGPT by addressing the following research questions (RQs):
\begin{itemize}[leftmargin=*, topsep=1pt]
	\item \textit{\textbf{RQ1.}} How does TempoGPT perform in time series reasoning tasks?
	\item \textit{\textbf{RQ2.}} Do TempoGPT's responses exhibit logical coherence?
	\item \textit{\textbf{RQ3.}} What impact do our innovative components have on TLMs in time series reasoning tasks?
	\item \textit{\textbf{RQ4.}} What other factors influence the performance of TLMs on time series reasoning tasks?
\end{itemize}

\subsection{Experimental Setup}

\subsubsection{\textbf{Dataset.}}
To explore TempoGPT's time series reasoning capabilities, we train it using the constructed multi-modal electrical dataset and evaluate it on five types of downstream tasks: 1) trend analysis; 2) trend forecast; 3) fault judgement; 4) fault diagnosis; 5) fault analysis. Among them, the first two tasks are mainly related to trend, primarily involving the analysis of time series; the latter three tasks are mainly related to reasoning, involving complex reasoning processes. Detailed descriptions of those data can be found in Section \ref{Multi-modal Data Generation}.

\subsubsection{\textbf{Evaluation.}}\label{sec:Evaluation Metrics}
To systematically evaluate TLMs' performance on time series reasoning tasks, we define three metrics to evaluate their capabilities:
\begin{itemize}[leftmargin=*, topsep=1pt]
	\item \textit{\textbf{Conclusion Accuracy (CA).}}  CA provides a convenient method to evaluate the time series reasoning capabilities by calculating the proportion of correct conclusions in the responses of TLMs.
	\item \textit{\textbf{Logical Reasoning Accuracy (LRA).}} Compared to CA, LRA offers a more rigorous evaluation. It only considers responses that accurately perceive time series information and derive correct conclusions from this perceived data as logically correct; all other responses are deemed incorrect.
    \item \textit{\textbf{Deception Rate (DR).}}  DR, which measures the proportion of responses that are conclusion-correct yet logically incorrect, serves as a complement to LRA.
\end{itemize}
We calculate CA through string matching, while LRA and DR are calculated via manual evaluation.
Among the three evaluation metrics mentioned, higher values for CA and LRA are preferable, whereas a lower DR is desirable. For more details about evaluation, please refer to Appendix \ref{apx:Evaluation Details}.

\subsubsection{\textbf{Baseline.}}
Depending on the method of representing time series data, our baseline methods are divided into two categories: 
\begin{itemize}[leftmargin=*, topsep=1pt]
	\item \textit{\textbf{Based on Textual Prompt.}}  We choose \textbf{GPT-3.5} \cite{GPT-3.5} and \textbf{GPT-4} \cite{GPT-4} as representatives of this approach, which directly converts time series data into textual prompt for input.
	\item \textit{\textbf{Based on Continuous Embedding.}} These methods represent temporal information within a continuous embedding space, thereby aligning temporal and textual information. Due to the current limitations of research on TLMs in time series reasoning tasks, we adopt the time series encoding methods of advanced TLMs as our baseline, including  \textbf{1) Linear-based.}  Encoding time series data with linear layers (e.g., GPT4MTS \cite{jia2024gpt4mts}); \textbf{2) Attention-based.} Encoding time series data with multi-head self-attention (e.g., Time-LLM \cite{jin2023time}); \textbf{3) MLP-based.} Encoding time series data with Multilayer Perceptron (MLP) (e.g., ChatTS \cite{xie2024chatts}).
\end{itemize}

\subsubsection{\textbf{Implementation.}}
We train all models in 4 $\times$ V100 GPUs. For TempoGPT based on GPT-2, in the pre-training stage, models are trained for approximately 6.6K steps with a learning rate of 1e-3 and a batch size of 24, which takes 1 hour; In the fine-tuning stage, the models are trained for approximately 7.2K steps with a learning rate of 1e-4 and a batch size of 12, which takes 30 minutes. More details of the training can be found in Appendix \ref{apx:Implementation Details}.

\subsection{Time Series Reasoning Performance (RQ1)} \label{sec: RQ1}
\begin{table*}[t]
\centering
\caption{Conclusion accuracy (\%) comparison of different temporal information representation methods in five types of time series reasoning tasks. During the fine-tuning stage, we use full fine-tuning for models based on GPT-2 and LoRA fine-tuning for other models. The values marked with \textcolor{red}{red} and \textcolor{blue}{blue} represent the best and second-best performance, respectively, for each corresponding task. For more experimental results, please refer to the Appendix \ref{apx:Time Series Reasoning}}
\label{tab: RQ1}
	\begin{tabular}{c|cc|cccc|cc}
		\toprule
		\multicolumn{1}{c|}{Method} & \multicolumn{2}{c|}{Textual Prompt}                                                                  & \multicolumn{4}{c|}{Continuous Embedding}                                                                                                                               & \multicolumn{2}{c}{\textbf{Discrete Embedding}}                                             \\
		\cmidrule(lr){2-3}\cmidrule(lr){4-7}\cmidrule(lr){8-9}
		
		\multirow{2}{*}{Model}      & \multicolumn{1}{c}{\multirow{2}{*}{GPT-3.5}} & \multicolumn{1}{c|}{\multirow{2}{*}{GPT-4}} & \multicolumn{1}{c}{GPT-2}  & \multicolumn{1}{c}{GPT-2} & \multicolumn{1}{c}{GPT-2}     & \multicolumn{1}{c|}{LLaMA-3.2-3B} & \multicolumn{1}{c}{\textbf{TempoGPT}} & \multicolumn{1}{c}{\textbf{TempoGPT}}     \\
		                             & \multicolumn{1}{c}{}                         & \multicolumn{1}{c|}{}                       & \multicolumn{1}{c}{(Linear)} & \multicolumn{1}{c}{(MLP)}   & \multicolumn{1}{c}{(Attention)} & \multicolumn{1}{c|}{(Linear)}           & \multicolumn{1}{c}{\textbf{(GPT-2)}}    & \multicolumn{1}{c}{\textbf{(LLaMA-3.2-3B)}} \\ \toprule
		\multicolumn{1}{l|}{Trend Analysis}                                       & 29.8                                         & 43.5                                        & \textcolor{red}{\textbf{99.3} }                      & 94.4                      & 93.7                          & 88.1                                                         & \textcolor{blue}{\textbf{95.8}}                         & 93.7                             \\
		\multicolumn{1}{l|}{Trend Forecast}                                       & 31.9                                         & 52.1                                        &\textcolor{blue}{\textbf{95.2}}                      & 92.0                                               & 87.2                             & 93.6                              & \textcolor{red}{\textbf{96.0}}                         & \textcolor{blue}{\textbf{95.2}}                             \\
		\multicolumn{1}{l|}{Fault Judgement}                                      & \textbackslash{}                             & \textbackslash{}                            & \textcolor{red}{\textbf{69.7}}                       & \textcolor{blue}{\textbf{68.4}}                                               & 42.1                             & 64.5                              & 65.8                         & 65.8                             \\
		\multicolumn{1}{l|}{Fault Diagnosis}                                      & \textbackslash{}                             & \textbackslash{}                            & 50.6                       & 63.6                      & 44.2                          & 29.9                                                         & \textcolor{blue}{\textbf{77.9}}                         & \textcolor{red}{\textbf{80.5}}                             \\
		\multicolumn{1}{l|}{Fault Analysis}                                       & \textbackslash{}                             & \textbackslash{}                            & 42.0                       & 37.0                      & 34.6                          & 16.0                                                       & \textcolor{blue}{\textbf{76.5}}                         & \textcolor{red}{\textbf{81.5}}                             \\ \midrule
		\multicolumn{1}{l|}{Avg. CA}                                              & 30.8                                         & 47.8                                        & 71.4                       & 71.1                      & 66.1                          & 52.7                                                         & \textcolor{blue}{\textbf{82.4}}                         & \textcolor{red}{\textbf{83.3}}                             \\ \bottomrule
	\end{tabular}
\end{table*}

As shown in Table \ref{tab: RQ1}, to evaluate the performance of TLMs in time series reasoning, we test these models across five types of time series reasoning tasks, comprising a total of 500 test questions, and evaluate them by conclusion accuracy.

For methods based on textual prompts, owing to the challenges posed by multiple variables and extensive redundant tokens, both GPT-3.5 and GPT-4 exhibit suboptimal performance in trend analysis and trend forecast tasks. We did not evaluate GPT-3.5 and GPT-4 on the remaining tasks, as LLMs face difficulties in zero-shot handling of relationships between temporal information and physical systems.
For TempoGPT, compare to TLMs based on continuous embedding, we only quantize continuous temporal embeddings to achieve consistent representation between textual and temporal information. As a result, regardless of the base model employed, the performance of TempoGPT surpasses TLMs based on continuous embedding.

Notably, tasks related to trends focus more on mining information from time series, while tasks involving temporal information and physical systems involve complex reasoning processes. Although TLMs based on continuous embedding excellently handle the fundamental time series analysis tasks related to trend, they perform poorly when dealing with the relationships between variables and physical systems.
This indicates ineffective alignment between temporal information and textual information.
Therefore, quantizing temporal embeddings to achieve a consistent representation of temporal and textual information enables better multi-modal alignment for TLMs, thereby enhancing the time series reasoning capabilities of TLMs.

\subsection{Study of Logical Reasoning (RQ2)}
Merely determining the correctness of conclusion is insufficient for effectively evaluating the reasoning capabilities of TLMs. Therefore, in this section, we systematically evaluate TLMs from both qualitative and quantitative perspectives to  determine whether they accurately perceive temporal information and logically reason to draw correct conclusions.
\subsubsection{\textbf{Qualitative Evaluation.}} \label{sec: Qualitative analysis}
\begin{figure*}[ht]
	\centering
	\includegraphics[width=\linewidth]{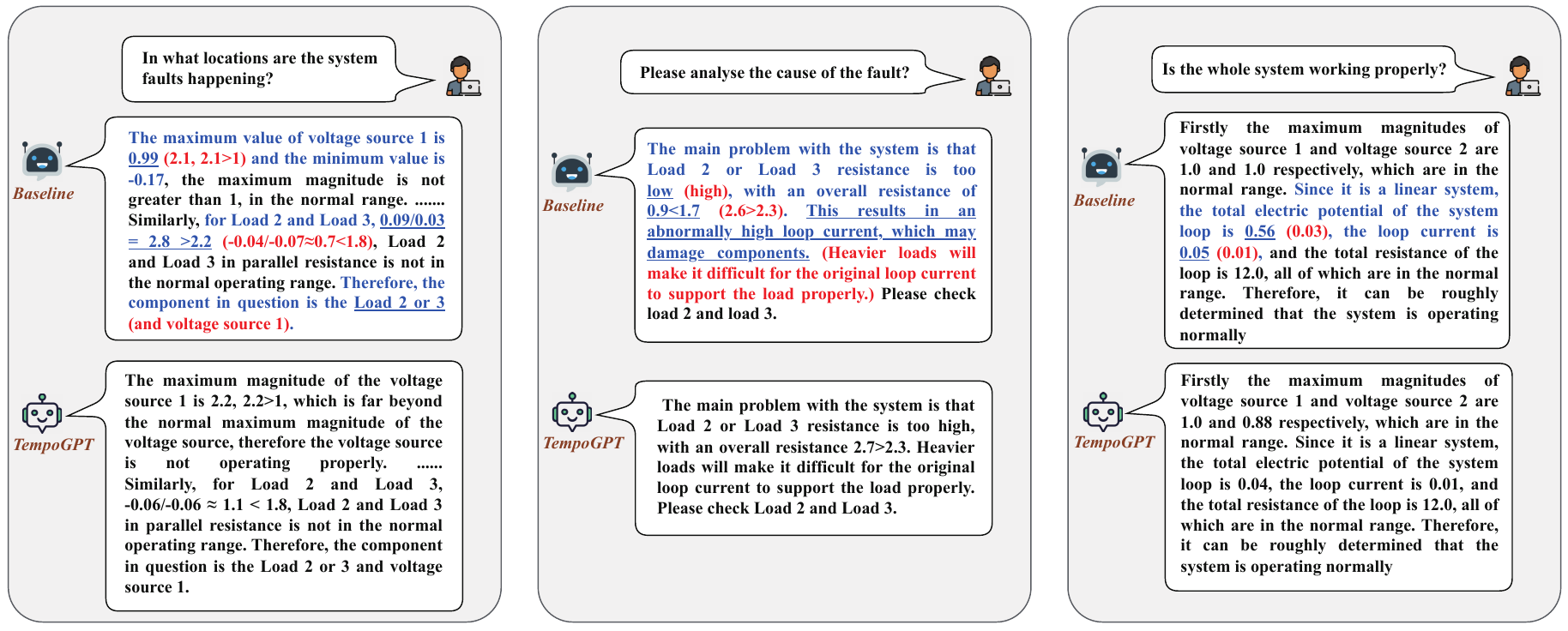}
	\caption{Quiz showcases of TempoGPT (GPT-2) and Baseline (GPT-2 based on linear) in the time series reasoning tasks. Responses highlighted in {\color{red}\textbf{red}} and {\color{blue}\textbf{blue}} indicate the true labels and incorrect perception or reasoning, respectively. For more response examples from TempoGPT, please refer to the Appendix \ref{apx: data}.}
	\label{fig:tempogpt_vs_baseline_qa}
\end{figure*}
As shown in Figure \ref{fig:tempogpt_vs_baseline_qa},  we list some examples of time series reasoning quiz for TempoGPT (GPT-2) and Baseline (GPT-2 based on linear), the best-performing method based on  GPT-2 and continuous embedding in Section \ref{sec: RQ1}. We can draw attention to the following two observations through interpreting and comparing these responses: 
\textbf{1) Temporal Information Perception.} The model must perceive the provided time series data. However, the baseline's perception of the time series data often differs significantly from the actual values. Although quantizing time series inevitably results in some loss of temporal information for TempoGPT, it still delivers a more precise perception of temporal information, closely approximating the actual values.
\textbf{2) Logical Reasoning.} The perception and logical reasoning capabilities influence the time series reasoning capabilities of TLMs. As shown on the left side of Figure \ref{fig:tempogpt_vs_baseline_qa}, the baseline's incorrect perception of the voltage value for voltage source 1 leads to an incorrect conclusion. Additionally, as the diagram in the middle and right of Figure \ref{fig:tempogpt_vs_baseline_qa} illustrates,  although the baseline model generates a correct conclusion,  it incorrectly perceives the temporal information, and the logic is inconsistent. Conversely, TempoGPT perceives data with greater accuracy and uses this information to reason logically and draw correct conclusions.

\subsubsection{\textbf{Quantitative evaluation.}} 
\begin{table*}[ht]
	\centering
	\caption{Manual quantitative evaluations for LRA (\%) and DR (\%) are presented using 25 test samples per task, where higher LRA and lower DR values are preferable. The values marked with \textcolor{red}{red} and \textcolor{blue}{blue} represent the best and second-best performance, respectively, for each corresponding task and metric. For more experimental results, please refer to the Appendix \ref{apx:Logical Reasoning Evaluation}}
	\label{tab: reasoning_analysis}
	\begin{tabular}{cl|cccccccc|cccc}
		\toprule
		\multicolumn{2}{c|}{\multirow{2}{*}{Model}} & \multicolumn{2}{c}{GPT-2}    & \multicolumn{2}{c}{GPT-2} & \multicolumn{2}{c}{GPT-2}       & \multicolumn{2}{c|}{LLaMA-3.2-3B} & \multicolumn{2}{c}{\textbf{TempoGPT}} & \multicolumn{2}{c}{\textbf{TempoGPT}}       \\
		\multicolumn{2}{c|}{}                       & \multicolumn{2}{c}{(Linear)} & \multicolumn{2}{c}{(MLP)} & \multicolumn{2}{c}{(Attention)} & \multicolumn{2}{c|}{(Linear)}     & \multicolumn{2}{c}{\textbf{(GPT-2)}}  & \multicolumn{2}{c}{\textbf{(LLaMA-3.2-3B)}} \\
		
		\cmidrule(lr){3-4} \cmidrule(lr){5-6} \cmidrule(lr){7-8} \cmidrule(lr){9-10}
		 \cmidrule(lr){11-12}  \cmidrule(lr){13-14}
		\multicolumn{2}{c|}{Metric}                 & LRA           & DR           & LRA         & DR          & LRA            & DR             & LRA             & DR              & LRA           & DR           & LRA              & DR              \\ \toprule
\multicolumn{2}{l|}{Fault Judgement}        & \textcolor{red}{\textbf{60.0 }}      & \textcolor{blue}{\textbf{4.0}}       & \textcolor{blue}{\textbf{56.0}}      & \textcolor{blue}{\textbf{4.0 }}      & \textcolor{blue}{\textbf{56.0 }}        & \textcolor{blue}{\textbf{4.0 }}         & 32.0          & \textcolor{red}{\textbf{0.0}}          & 48.0        & 8.0       & 48.0          & \textcolor{blue}{\textbf{4.0}}           \\
\multicolumn{2}{l|}{Fault Diagnosis}        & 44.0       & 16.0      & 48.0      & 24.0      & 28.0         & 24.0         & 24.0          & 12.0          & \textcolor{blue}{\textbf{76.0}}        & \textcolor{red}{\textbf{0.0}}        & \textcolor{red}{\textbf{84.0}}         & \textcolor{blue}{\textbf{4.0}}           \\
\multicolumn{2}{l|}{Fault Analysis}         & 52.0        & 20.0       & 36.0      & 20.0      & 44.0         & 20.0        & 16.0          & 12.0          & \textcolor{blue}{\textbf{68.0}}        &\textcolor{red}{\textbf{ 0.0 }}       & \textcolor{red}{\textbf{76.0}}          &\textcolor{red}{\textbf{0.0}}           \\ \midrule
\multicolumn{2}{l|}{Average}                & 52.0        & 13.3       & 46.7      & 16.0      & 42.7         & 16.0        & 24.0          & 8.0           & \textcolor{blue}{\textbf{64.0}}        & \textcolor{red}{\textbf{2.7}}     & \textcolor{red}{\textbf{69.3}}          &\textcolor{red}{\textbf{2.7}}           \\ \bottomrule

	\end{tabular}
\end{table*}

As tasks related to trends primarily concentrate on extracting information from time series rather than engaging in complex reasoning processes, we select the remaining tasks for quantitative evaluation.  We randomly choose 25 test samples for each task, totaling 75 samples, and evaluate the responses of TLMs by LRA and DR. The statistical results for LRA and DR are shown in Table  \ref{tab: reasoning_analysis}. 

TempoGPT considerably surpasses all TLMs based on continuous embeddings in terms of average LRA, demonstrating its superior ability to derive correct conclusions through its inherent logical reasoning capabilities. Furthermore, TempoGPT exhibits low DR values, indicating a reduced likelihood of random guessing and a stronger reliance on logical reasoning to provide accurate responses. In contrast, TLMs based on continuous embedding show DR values exceeding 20\% in certain tasks, suggesting that even if their conclusions are correct, there is a high probability that their responses are logically inconsistent. As a result, TempoGPT more effectively aligns temporal and textual information, harnessing the reasoning capabilities of LLMs to tackle complex time series reasoning tasks.

\subsection{Ablation Study (RQ3)}
In this section, we primarily investigate whether the quantization component and pre-training have a positive effect on the time series reasoning capabilities of TLMs.

To investigate the influence of quantization on the time series reasoning capabilities of TLMs, we train VQ-VAE using linear, self-attention, and MLP as core components to generate the corresponding temporal codebook. Subsequently, TLMs can tokenize temporal embeddings into discrete tokens using this temporal codebook. We also expand the embedding matrix to accommodate these temporal tokens, thus completing the quantization of the time series data. Since TLMs perform well on trend-related tasks, we only test the TLMs before and after adding quantization on three reasoning-related tasks. The experimental results are shown in Table \ref{tab: Ablation Study}. With few exceptions, it is evident that the introduction of quantization generally yields significant improvements for nearly all TLMs based on continuous embeddings, with some metrics exhibiting improvement effects surpassing 100\%.  This demonstrates the effectiveness of the quantization component, showing that quantizing temporal embeddings to achieve consist representation pattern with textual information can significantly enhance the time series reasoning capabilities of TLMs.

\begin{figure}[ht]
	\centering
	\begin{subfigure}[b]{0.45\linewidth}
		\includegraphics[height=\linewidth]{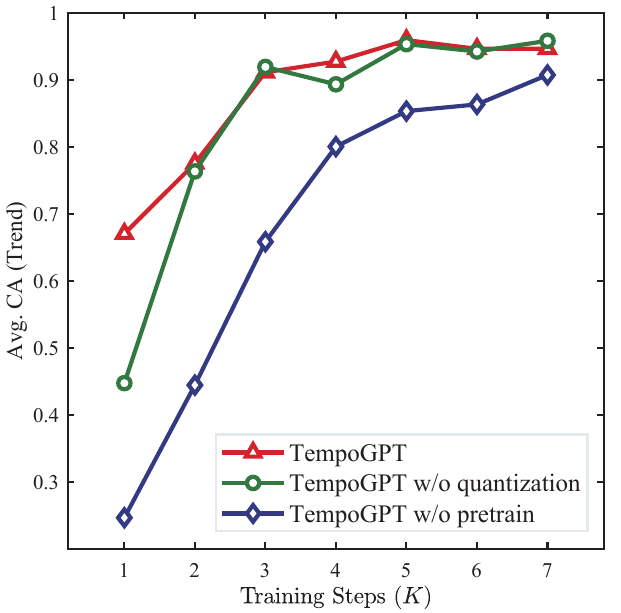}
		\caption{Trend.}
		\label{fig:RQ3_trend}
	\end{subfigure}
	\hfill
	\begin{subfigure}[b]{0.45\linewidth}
		\includegraphics[height=\linewidth]{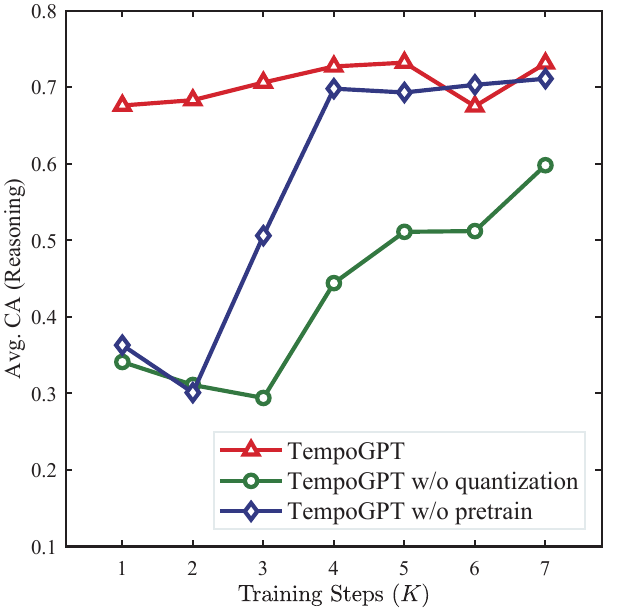}
		\caption{Reasoning.}
		\label{fig:RQ3_reason}
	\end{subfigure}
	\caption{The CA of TempoGPT and its variants in tasks related to trend and reasoning during the fine-tuning stage. For more experimental results, please refer to Appendix \ref{apx: Ablation Study}.}
	\label{fig:RQ3_trend_reason}
\end{figure}

Meanwhile, to further explore the impact of pre-training and quantization on the time series reasoning capabilities of TLMs, as shown in the Figure \ref{fig:RQ3_trend_reason}, we present CA in tasks related to trend and reasoning during the fine-tuning stage for TempoGPT (GPT2) and its two variants. 
After removing pre-training and quantization, the convergence speeds of the TLMs on both types of tasks have significantly slowed down. Particularly for tasks related to trends, the convergence speeds and effectiveness are notably inferior after the removal of pre-training compared to the other two scenarios, indicating that pre-training enables TLMs to learn fundamental temporal information. 
Moreover, even though the corresponding variant can eventually achieve results comparable to TempoGPT on tasks related to trends after removing quantization, it encounter difficulties with tasks involving reasoning. This observation suggests that quantization can effectively enhance multi-modal alignment and bolster the reasoning capabilities of TLMs.

\begin{table*}[t]
	\centering
	\caption{Performance promotion obtained by our quantization component. We present the advancements for five types of TLMs based on continuous embedding, before and after the incorporation of a quantization component. These improvements are reported across three tasks related to reasoning, with metrics including CA (\%), LRA (\%), and DR (\%).}
	\label{tab: Ablation Study}
	\resizebox{\textwidth}{!}{
		\begin{tabular}{cl|ccccccccccccccc}
			\toprule
			\multicolumn{2}{c|}{\multirow{2}{*}{Model}}                                      & \multicolumn{3}{c}{GPT-2}                                                & \multicolumn{3}{c}{GPT-2}                                                & \multicolumn{3}{c}{GPT-2}                                                & \multicolumn{3}{c}{LLaMA-3.2-1B}                                           & \multicolumn{3}{c}{LLaMA-3.2-3B}                    \\
			\multicolumn{2}{c|}{}                                                            & \multicolumn{3}{c}{(Linear)}                                         & \multicolumn{3}{c}{(MLP)}                                              & \multicolumn{3}{c}{(Attention)}                                        & \multicolumn{3}{c}{(Linear)}                                            & \multicolumn{3}{c}{(Linear)}                       \\ 
			\cmidrule(lr){3-5} \cmidrule(lr){6-8} \cmidrule(lr){9-11} \cmidrule(lr){12-14} \cmidrule(lr){15-17} 
			\multicolumn{2}{c|}{Metric}                                                      & CA              & LRA             & \multicolumn{1}{c|}{DR}              & CA              & LRA             & \multicolumn{1}{c|}{DR}              & CA              & LRA             & \multicolumn{1}{c|}{DR}              & CA               & LRA              & \multicolumn{1}{c|}{DR}              & CA               & LRA              & DR            \\ \toprule
			\multicolumn{1}{c|}{\multirow{2}{*}{\begin{tabular}[1]{@{}c@{}}Fault\\Judgement\end{tabular}}} & Original               & 69.7            & 60.0            & \multicolumn{1}{c|}{4.0}             & 68.4            & 56.0            & \multicolumn{1}{c|}{4.0}             & 64.5            & 56.0            & \multicolumn{1}{c|}{4.0}             & 39.5             & 28.0             & \multicolumn{1}{c|}{0.0}             & 42.1             & 32.0             & 0.0           \\
			\multicolumn{1}{c|}{}                                   & \textbf{+Quantization} & 61.8            & 52.0            & \multicolumn{1}{c|}{\textbf{0.0}}    & \textbf{84.2}   & \textbf{68.0}   & \multicolumn{1}{c|}{8.0}             & \textbf{69.7}   & \textbf{60.0}   & \multicolumn{1}{c|}{\textbf{4.0}}    & \textbf{65.8}    & \textbf{56.0}    & \multicolumn{1}{c|}{\textbf{0.0}}    & \textbf{63.2}    & \textbf{48.0}    & \textbf{0.0}  \\ \midrule
			\multicolumn{1}{c|}{\multirow{2}{*}{\begin{tabular}[1]{@{}c@{}}Fault\\Diagnosis\end{tabular}}} & Original               & 50.6            & 44.0            & \multicolumn{1}{c|}{16.0}            & 63.6            & 48.0            & \multicolumn{1}{c|}{24.0}            & 44.2            & 28.0            & \multicolumn{1}{c|}{24.0}            & 40.3             & 32.0             & \multicolumn{1}{c|}{20.0}            & 29.9             & 24.0             & 12.0          \\
			\multicolumn{1}{c|}{}                                   & \textbf{+Quantization} & \textbf{77.9}   & \textbf{68.0}   & \multicolumn{1}{c|}{\textbf{16.0}}   & \textbf{79.2}   & \textbf{84.0}   & \multicolumn{1}{c|}{\textbf{4.0}}    & \textbf{81.8}   & \textbf{84.0}   & \multicolumn{1}{c|}{\textbf{4.0}}    & \textbf{75.3}    & \textbf{64.0}    & \multicolumn{1}{c|}{\textbf{12.0}}   & \textbf{79.2}    & \textbf{60.0}    & 24.0          \\ \midrule
			\multicolumn{1}{c|}{\multirow{2}{*}{\begin{tabular}[1]{@{}c@{}}Fault\\Analysis\end{tabular}}}  & Original               & 42.0            & 52.0            & \multicolumn{1}{c|}{20.0}            & 37.0            & 36.0            & \multicolumn{1}{c|}{20.0}            & 34.6            & 44.0            & \multicolumn{1}{c|}{20.0}            & 14.8             & 8.0              & \multicolumn{1}{c|}{20.0}            & 16.0             & 16.0             & 12.0          \\
			\multicolumn{1}{c|}{}                                   & \textbf{+Quantization} & \textbf{75.3}   & \textbf{76.0}   & \multicolumn{1}{c|}{\textbf{8.0}}    & \textbf{85.2}   & \textbf{76.0}   & \multicolumn{1}{c|}{\textbf{0.0}}    & \textbf{88.9}   & \textbf{92.0}   & \multicolumn{1}{c|}{\textbf{0.0}}    & \textbf{65.4}    & \textbf{60.0}    & \multicolumn{1}{c|}{\textbf{16.0}}   & \textbf{65.4}    & \textbf{56.0}    & \textbf{12.0} \\ \midrule
			\multicolumn{1}{c|}{\multirow{3}{*}{Average}}           & Original               & 54.1            & 52.0            & \multicolumn{1}{c|}{13.3}            & 56.4            & 46.7            & \multicolumn{1}{c|}{16.0}            & 47.7            & 42.7            & \multicolumn{1}{c|}{16.0}            & 31.5             & 22.7             & \multicolumn{1}{c|}{13.3}            & 29.3             & 24.0             & 8.0           \\
			\multicolumn{1}{c|}{}                                   & \textbf{+Quantization} & \textbf{71.7}   & \textbf{65.3}   & \multicolumn{1}{c|}{\textbf{8.0}}    & \textbf{82.9}   & \textbf{76.0}   & \multicolumn{1}{c|}{\textbf{4.0}}    & \textbf{80.1}   & \textbf{78.7}   & \multicolumn{1}{c|}{\textbf{2.7}}    & \textbf{68.8}    & \textbf{60.0}    & \multicolumn{1}{c|}{\textbf{9.3}}    & \textbf{69.3}    & \textbf{54.7}    & 12.0          \\
			\cmidrule(lr){2-17} 
			
			\multicolumn{1}{c|}{}                                   & \textbf{Promotion }             & \textbf{32.5\%} & \textbf{25.6\%} & \multicolumn{1}{c|}{\textbf{40.0\%}} & \textbf{47.0\%} & \textbf{62.9\%} & \multicolumn{1}{c|}{\textbf{75.0\%}} & \textbf{67.9\%} & \textbf{84.4\%} & \multicolumn{1}{c|}{\textbf{83.3\%}} & \textbf{118.5\%} & \textbf{164.7\%} & \multicolumn{1}{c|}{\textbf{30.0\%}} & \textbf{136.1\%} & \textbf{127.8\%} & -50.0\%       \\ \bottomrule
	\end{tabular}}
\end{table*}

\subsection{Model Analysis (RQ4)}
\begin{table}[ht]
	\centering
	\caption{The impact of different base models and fine-tuning methods on the time series reasoning capabilities of TempoGPT. We report the average values of CA (\%), LRA (\%), and DR (\%). For detailed experimental results, please refer to Appendix \ref{apx: more_result}.}
	\label{tab: Model Analysis}
	\begin{tabular}{ll|ccc|c}
		\toprule
		\multicolumn{1}{l}{\multirow{2}{*}{Base   Model}} & \multicolumn{1}{l|}{\multirow{2}{*}{Params}} & \multicolumn{3}{c|}{Metric} & \multicolumn{1}{c}{\multirow{2}{*}{Rank}} \\
		 \cmidrule(lr){3-5}
		\multicolumn{1}{c}{}                              & \multicolumn{1}{c|}{}                        & CA      & LRA     & DR      & \multicolumn{1}{c}{}                      \\ \midrule
		GPT-2 (LoRA)                                      & 125M                                         & 40.7    & 24.0    & 34.7    & 6                                         \\
		GPT-2 (Full)                                      & 125M                                         & 82.4    & 64.0    & 2.7     & 3                                         \\
		Tiny-llama                                        & 1.1B                                         & 82.0    & 68.0    & 6.7     & 5                                         \\
		LLaMA-3.2-1B                                      & 1.2B                                         & 79.3    & 57.3    & 5.3     & 4                                         \\
		Phi-2                                             & 2.7B                                         & 82.7    & 68.0    & 2.7     & 2                                         \\
		LLaMA-3.2-3B                                      & 3.2B                                         & 83.3    & 69.3    & 2.7     & 1                                         \\ \bottomrule
	\end{tabular}
\end{table}
To further explore the impact of different base models and fine-tuning methods on TempoGPT's time series reasoning capabilities, we studied the performance of TempoGPT in solving time series reasoning tasks using GPT-2 (Full), GPT-2 (LoRA), Tiny-llama, LLaMA-3.2-1B, Phi-2, and LLaMA-3.2-3B as base models. Among them, GPT-2 (Full) utilizes full fine-tuning during the fine-tuning stage, while other models use LoRA fine-tuning during the fine-tuning stage. 

As shown in Table \ref{tab: Model Analysis}, we report the average CA, LRA, and DR for different models in solving time series reasoning tasks. Furthermore, the rank is determined using an entropy-based TOPSIS method \cite{chen2021effects}. The analysis shows that, excluding GPT-2 (Full) and with uniform training methods, there is a clear tendency for larger models to achieve better rank. This trend holds true for CA, LRA, and DR, where, with few exceptions, performance improves as the size of base model increases. 
It is worth noting that for TLMs based on continuous embedding, Table \ref{tab: Ablation Study} presents a trend opposite to the aforementioned conclusion and related studies \cite{liu2024autotimes, jin2023time, liu2024visual}. This indicates that for TLMs based on continuous embedding, as the base model size increases, it may become more challenging to effectively leverage the LLMs' inherent reasoning capabilities.

In addition, the performance achieved by the GPT-2 (Full) is only lower to those using the two \textasciitilde3B base models. Full fine-tuning involves altering a large number of GPT-2's original parameters, which may compromise its inherent natural language processing capabilities. However, in specific downstream domains, applying full fine-tuning to lightweight models could possibly achieve results comparable to those of larger models in a more computationally efficient manner.

\section{Conclusion and Future Work}
In response to the challenges of multi-modal time series language models (TLMs) in complex reasoning tasks, we propose a multi-modal time series data construction method and design a multi-modal time series language model, TempoGPT. Specially,  we construct complex time series reasoning data by exploring the relationships between time series and physical systems within a white-box system, while also release an electrical multi-modal time series dataset.  Additionally, proposed 
TempoGPT tokenize time series into temporal tokens via quantization encoding, then employs a shared embedding layer to process both temporal and textual tokens, achieving a consistent representation of temporal and textual information.
Experimentally, TempoGPT exhibits robust perception and reasoning capabilities, achieving state-of-the-art in the constructed time series reasoning tasks. Under systematic analysis, TempoGPT demonstrates outstanding framework generality, effectively enhancing the multi-modal alignment and time series reasoning capabilities of TLMs through the quantization of temporal embeddings.
In the future, we will explore systematic time series reasoning benchmark, involving more complex time series data and challenging downstream tasks, requiring TLMs to have robust temporal perception and reasoning capabilities.

\begin{acks}
	This  work is supported by the the National Natural Science Foundation of China (Grant No. 62473383, No. 62394340) and the Young Elite Scientists Sponsorship Program by CAST (Grant No. 2023QNRC001).
\end{acks}

\clearpage

\bibliographystyle{ACM-Reference-Format}
\balance
\bibliography{online.bib}

\clearpage
\appendix

\section{Data details} \label{apx: data}
\begin{figure*}[hb]
	\centering
	\includegraphics[width=\linewidth]{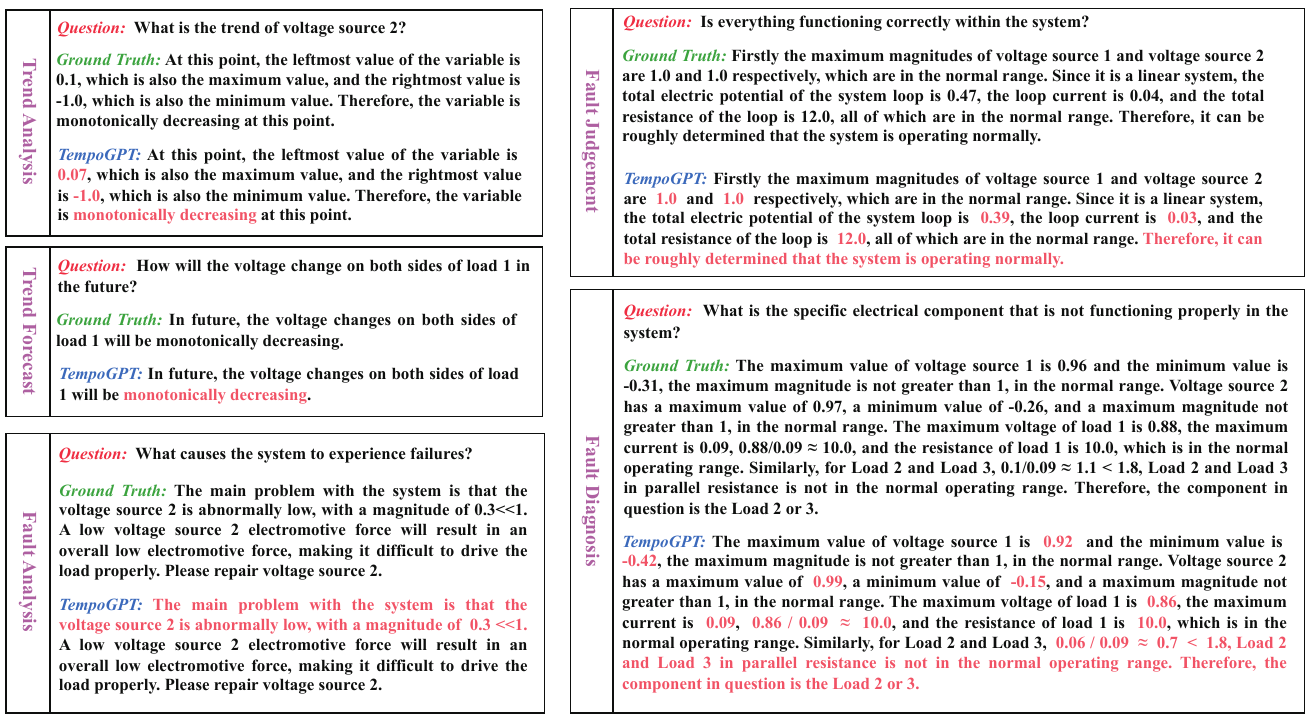}
	\caption{Display of data in the fine-tuning stage, as well as examples of responses from TempoGPT.}
	\label{fig: apx_QA}
\end{figure*}

Due to space limitations, we present the data from the fine-tuning stage in more detail here. This primarily includes five types of time series reasoning tasks: trend analysis, trend forecast, fault judgement, fault diagnosis, and fault analysis, with specific content as shown in Figure \ref{fig: apx_QA}. Meanwhile, we display related response examples from TempoGPT. TempoGPT is capable of precisely perceiving temporal information, reasoning logically, and correctly drawing conclusions.

\section{Training details}\label{apx:Training details}
\subsection{Evaluation Details}\label{apx:Evaluation Details}
We define three metrics (CA, LRA, and DR) to evaluate the model's capability to handle time series reasoning tasks. 

For the CA, we constructed a test set comprising five types of time series reasoning tasks, totaling 500 test samples. These test samples are not included in the training and validation sets. We evaluate the correctness of the model's conclusions by matching the strings of the model's responses with the strings of the reference conclusions, as the conclusions are concise and clear. The CA is calculated by estimating the number of correct responses for each task relative to the total number of samples for that task.

For LRA and DR, aside from the tasks related to trend (trend forecast and trend analysis tasks), we randomly select 25 test samples for each task from the aforementioned test set, totaling 75 samples. We manually evaluate whether the model's responses in terms of perception, reasoning, and conclusions are appropriate and correct. Specifically, for perception, we consider that the model's estimation of the statistical values of time series data should not deviate significantly from the real values. Meanwhile, the estimation of the state of components should be correct. For the reasoning process, if a model's conclusion is correct, then the information obtained from perception should logically lead to the correct conclusion. If not, it is considered a logical reasoning error. For the correctness of the model's conclusions, it is judged using the same method in evaluating CA. After the manual evaluation is completed, we calculate the values of LRA and DR based on the definitions provided in the section \ref{sec:Evaluation Metrics}.

\subsection{Implementation Details}\label{apx:Implementation Details}

We train all models with 4 $\times$ V100. According to \cite{liu2024visual}, epochs are set to 1 in the pre-training stage, involving about 6.6K training steps. The epochs are set to 2 in the fine-tuning stage, involving about 7.2K training steps.  We use a cosine learning rate, setting 1e-3 for the pre-training stage and 1e-4 for the fine-tuning stage.

For TLMs based on GPT-2, the batch size is set to 24 during the pre-training stage and reduced to 12 during the fine-tuning stage, with full fine-tuning implemented. For TLMs based on other base models, the batch size is consistently set to 4 throughout both the pre-training and fine-tuning stages, incorporating full fine-tuning during pre-training and LoRA fine-tuning during the fine-tuning stage. Furthermore, for LoRA fine-tuning, LoRA $r$, LoRA $\alpha$, LoRA dropout are set to 8, 16, 0.05, respectively.

\section{Extra Experimental Result} \label{apx: more_result}
\subsection{Time Series Reasoning} \label{apx:Time Series Reasoning}
\begin{table*}[hb]
	\centering
	\caption{Extra experimental result about conclusion accuracy (\%) comparison in five types of time series reasoning tasks. The values marked with \textcolor{red}{red} and \textcolor{blue}{blue} represent the best and second-best performance, respectively, for each corresponding task.}
	\label{tab: reamain result of RQ1}
	\begin{tabular}{cl|ccc|ccc}
		\toprule
		\multicolumn{2}{c|}{Method}                 & \multicolumn{3}{c|}{Continuous Embedding}  & \multicolumn{3}{c}{Discrete Embedding} \\ \cmidrule(lr){3-5} \cmidrule(lr){6-8}
		\multicolumn{2}{c|}{\multirow{2}{*}{Model}} & LLaMA-3.2-1B & LLaMA-3.2-1B & LLaMA-3.2-3B & TempoGPT      & TempoGPT    & TempoGPT \\
		\multicolumn{2}{l|}{}                       & (Linear)       & (Attention)    & (Attention)    & (LLaMA-3.2-1B)  & (Tiny-llama)  & (Phi-2)    \\ \midrule
		\multicolumn{2}{l|}{Trend Analysis}         & 92.3         & \textcolor{red}{\textbf{99.3}}         & \textcolor{red}{\textbf{99.3}}         & 95.8          & 93.7        & \textcolor{blue}{\textbf{97.2}}     \\
		\multicolumn{2}{l|}{Trend Forecast}         & 88.8         & \textcolor{red}{\textbf{96.8}}         & 93.6         & \textcolor{blue}{\textbf{96.0}}          & 95.2        & \textcolor{red}{\textbf{96.8}}     \\
		\multicolumn{2}{l|}{Fault Judgement}        & 39.5         & \textcolor{red}{\textbf{67.1}}         & 64.5         & \textcolor{blue}{\textbf{65.8}}          & \textcolor{blue}{\textbf{65.8}}        & \textcolor{red}{\textbf{67.1}}     \\
		\multicolumn{2}{l|}{Fault Diagnosis}        & 40.3         & 61.0         & 50.6         & 77.9          & \textcolor{blue}{\textbf{80.5}}        & \textcolor{red}{\textbf{87.0}}     \\
		\multicolumn{2}{l|}{Fault Analysis}         & 14.8         & 42.0         & 30.9         & \textcolor{blue}{\textbf{76.5}}          & \textcolor{red}{\textbf{81.5}}        & 65.4     \\ \midrule
		\multicolumn{2}{l|}{Avg. CA}                & 55.1         & 73.2         & 67.8         & 82.4          & \textcolor{red}{\textbf{83.3}}        & \textcolor{blue}{\textbf{82.7}}     \\ \bottomrule
	\end{tabular}
\end{table*}

In this section, we maintain the experimental setup consistent with Section \ref{sec: RQ1}, and provide additional relevant experimental results. This primarily involves using different base models and encoding methods. For specific details, please refer to Table \ref{tab: reamain result of RQ1}.
Regardless of the base model or encoder employed, within the tempoGPT framework, the performance in processing time series reasoning tasks significantly surpasses that of methods based on continuous embedding. This demonstrates that representing time series data in a discrete embedding space, forming the same representation pattern as textual information, can significantly improve the time series reasoning capabilities of TLMs.

\subsection{Logical Reasoning} \label{apx:Logical Reasoning Evaluation}
\begin{table*}[hb]
		\centering
		
	\caption{Extra experimental result about LRA (\%) and DR (\%) comparison in tasks related to reasoning, where higher LRA and lower DR values are preferable. The values marked with \textcolor{red}{red} and \textcolor{blue}{blue} represent the best and second-best performance, respectively, for each corresponding task and metric.}
	\label{tab: reamain result of RQ2}
	\begin{tabular}{cc|cccccc|cccccc}
		\toprule
		\multicolumn{2}{c|}{\multirow{2}{*}{Model}} & \multicolumn{2}{c}{LLaMA-3.2-1B} & \multicolumn{2}{c}{LLaMA-3.2-1B} & \multicolumn{2}{c|}{LLaMA-3.2-3B} & \multicolumn{2}{c}{TempoGPT}     & \multicolumn{2}{c}{TempoGPT}   & \multicolumn{2}{c}{TempoGPT} \\
		\multicolumn{2}{c|}{}                       & \multicolumn{2}{c}{(Linear)}   & \multicolumn{2}{c}{(Attention)}& \multicolumn{2}{c|}{(Attention)} & \multicolumn{2}{c}{(LLaMA-3.2-1B)} & \multicolumn{2}{c}{(Tiny-llama)} & \multicolumn{2}{c}{(Phi-2)}    \\
		\cmidrule(lr){3-4} \cmidrule(lr){5-6} \cmidrule(lr){7-8} \cmidrule(lr){9-10}
		\cmidrule(lr){11-12}  \cmidrule(lr){13-14}
		\multicolumn{2}{c|}{Metric}                 & LRA             & DR             & LRA             & DR             & LRA             & DR              & LRA             & DR             & LRA            & DR            & LRA           & DR           \\ \midrule
\multicolumn{2}{l|}{Fault Judgement}        & 28.0            & \textcolor{red}{\textbf{0.0}}           & 44.0            & 12.0           & 44.0            & 12.0            & 48.0            & \textcolor{red}{\textbf{0.0}}            & \textcolor{red}{\textbf{56.0}}           & 8.0           & \textcolor{blue}{\textbf{52.0}}          & \textcolor{blue}{\textbf{4.0}}          \\
\multicolumn{2}{l|}{Fault Diagnosis}        & 32.0            & 20.0           & 52.0            & \textcolor{blue}{\textbf{8.0}}            & 32.0            & 24.0            & 52.0            & 16.0           & \textcolor{blue}{\textbf{72.0}}           & 12.0          & \textcolor{red}{\textbf{76.0}}          & \textcolor{red}{\textbf{4.0}}          \\
\multicolumn{2}{l|}{Fault Analysis}         & 8.0             & 20.0           & 48.0            & \textcolor{blue}{\textbf{8.0}}            & 32.0            & 16.0            & \textcolor{blue}{\textbf{72.0}}            & \textcolor{red}{\textbf{0.0}}            & \textcolor{red}{\textbf{76.0}}           & \textcolor{red}{\textbf{0.0}}           & \textcolor{red}{\textbf{76.0}}          & \textcolor{red}{\textbf{0.0}}          \\  \midrule
\multicolumn{2}{l|}{Average}                & 22.7            & 13.3           & 48.0            & 9.3            & 36.0            & 17.3            & \textcolor{blue}{\textbf{57.3}}            & \textcolor{blue}{\textbf{5.3}}            & \textcolor{red}{\textbf{68.0}}           & 6.7           & \textcolor{red}{\textbf{68.0}}          & \textcolor{red}{\textbf{2.7}}          \\ \bottomrule
	\end{tabular}
\end{table*}

In this section, consistent with the setup in Section \ref{sec: Qualitative analysis}, we present more experimental results regarding the LRA and DR metrics. The specific details are shown in Table \ref{tab: reamain result of RQ2}. Comparing with Table \ref{tab: reasoning_analysis}, it is clearly that when the parameters of the base model increases, the performance under the TempoGPT framework is significantly better than that of TLMs based on continuous embedding. This highlights the inherent challenge faced by TLMs based on continuous embedding in effectively harnessing the reasoning capabilities of LLMs, particularly when the model size is substantial. Moreover, TempoGPT significantly outperforms TLMs based on continuous embedding on both  LRA and DR. This suggests that quantizing the temporal embeddings to achieve a consistent representation pattern with textual information can enhance the multi-modal alignment  and reasoning capabilities of TLMs.

\subsection{Ablation Study} \label{apx: Ablation Study}
We report the extended experiments and the detailed data for Figure \ref{fig:RQ3_trend_reason} in Table \ref{tab: reamain result of RQ3} and Table \ref{tab: reamain result of RQ3_2}, respectively. 
Both Table \ref{tab: reamain result of RQ3} and Table \ref{tab: reamain result of RQ3_2} show that after removing pre-training, the performance of TLMs deteriorates on tasks related to trend. Additionally, one  observation can be made from Table \ref{tab: reamain result of RQ3}: using LoRA for fine-tuning significantly reduces the capability of TLMs to solve time series reasoning tasks. From Table  \ref{tab: reamain result of RQ3_2}, it can also be observed that after removing quantization, the performance of TLMs significantly deteriorates on tasks related to reasoning; regardless of whether pre-training or quantization is removed, the convergence speed of TLMs during training becomes significantly slower.
In summary, both quantization and pre-training accelerate the convergence speed. Pre-training can improve TLMs' processing of temporal information, while quantization significantly enhances multi-modal alignment and TLMs' reasoning capabilities.

\begin{table*}[hb]
	\centering
	\caption{Ablation study results for pre-training and fine-tuning methods under the TempoGPT framework, reporting CA, LRA, and DR metrics.}
	\label{tab: reamain result of RQ3}

	\resizebox{\textwidth}{!}{
	\begin{tabular}{c|l|ccc|cc|cc|cc|cc}
		\toprule
		\multicolumn{1}{l|}{Metric} & \multicolumn{1}{l|}{Task} & GPT-2 & w/o Pretrain & w/o Full & LLaMA-3.2-1B & w/o Pretrain & LLaMA-3.2-3B & w/o Pretrain & Tiny-llama & w/o Pretrain & Phi-2 & w/o Pretrain \\ \midrule
		\multirow{6}{*}{CA}         & Trend  Analysis           & 95.8  & 90.2         & 33.6     & 92.3         & 95.1         & 93.7         & 90.2         & 93.0       & 82.5         & 97.2  & 94.4         \\
		& Trend Forecast            & 96.0  & 93.6         & 32.0     & 94.4         & 96.0         & 95.2         & 96.8         & 94.4       & 92.0         & 96.8  & 96.0         \\
		& Fault Judgement           & 65.8  & 56.6         & 61.8     & 56.6         & 61.8         & 65.8         & 57.9         & 65.8       & 63.2         & 67.1  & 71.1         \\
		& Fault Diagnosis           & 77.9  & 74.0         & 45.5     & 75.3         & 77.9         & 80.5         & 77.9         & 77.9       & 77.9         & 87.0  & 77.9         \\
		& Fault Analysis            & 76.5  & 61.7         & 30.9     & 77.8         & 76.5         & 81.5         & 76.5         & 79.0       & 75.3         & 65.4  & 79.0         \\ \cmidrule(lr){2-13} 
		& Average                     & 82.4  & 75.2         & 40.7     & 79.3         & 81.5         & 83.3         & 79.9         & 82.0       & 78.2         & 82.7  & 83.7         \\ \midrule
		\multirow{4}{*}{LRA}        & Fault Judgement           & 48.0  & 40.0         & 24.0     & 48.0         & 52.0         & 48.0         & 52.0         & 56.0       & 52.0         & 52.0  & 64.0         \\
		& Fault Diagnosis           & 76.0  & 52.0         & 12.0     & 52.0         & 72.0         & 84.0         & 68.0         & 72.0       & 68.0         & 76.0  & 72.0         \\
		& Fault Analysis            & 68.0  & 60.0         & 36.0     & 72.0         & 64.0         & 76.0         & 84.0         & 76.0       & 72.0         & 76.0  & 76.0         \\ \cmidrule(lr){2-13} 
		& Average                      & 64.0  & 50.7         & 24.0     & 57.3         & 62.7         & 69.3         & 68.0         & 68.0       & 64.0         & 68.0  & 70.7         \\ \midrule
		\multirow{4}{*}{DR}         & Fault Judgement           & 8.0   & 0.0          & 28.0     & 0.0          & 0.0          & 4.0          & 4.0          & 8.0        & 0.0          & 4.0   & 4.0          \\
		& Fault Diagnosis           & 0.0   & 32.0         & 52.0     & 16.0         & 8.0          & 4.0          & 8.0          & 12.0       & 12.0         & 4.0   & 4.0          \\
		& Fault Analysis            & 0.0   & 12.0         & 24.0     & 0.0          & 0.0          & 0.0          & 0.0          & 0.0        & 0.0          & 0.0   & 0.0          \\ \cmidrule(lr){2-13} 
		& Average                     & 2.7   & 14.7         & 34.7     & 5.3          & 2.7          & 2.7          & 4.0          & 6.7        & 4.0          & 2.7   & 2.7          \\ \bottomrule
	\end{tabular}}
\end{table*}

\begin{table*}[hb]
		\centering
	\caption{The detail concept about CA (\%) of TempoGPT and its variants in time series reasoning tasks during the fine-tuning stage. After removing quantization, the performance of TLMs significantly deteriorates on tasks related to reasoning. After removing pre-training, the performance significantly deteriorates on tasks related to trend.}
	\label{tab: reamain result of RQ3_2}
	\resizebox{\textwidth}{!}{
	\begin{tabular}{l|c|ccccc|ccc}
		\toprule
\multicolumn{1}{l|}{Model}                    & Steps (K) & \multicolumn{1}{c}{\begin{tabular}[c]{@{}c@{}}Trend\\ Analysis\end{tabular}} & \multicolumn{1}{c}{\begin{tabular}[c]{@{}c@{}}Trend\\ Forecast\end{tabular}} & \multicolumn{1}{c}{\begin{tabular}[c]{@{}c@{}}Fault\\ Judgement\end{tabular}} & \multicolumn{1}{c}{\begin{tabular}[c]{@{}c@{}}Fault\\ Diagnosis\end{tabular}} & \multicolumn{1}{c|}{\begin{tabular}[c]{@{}c@{}}Fault\\ Analysis\end{tabular}} & \multicolumn{1}{c}{Avg. CA} & \multicolumn{1}{c}{\begin{tabular}[c]{@{}c@{}}Avg. CA\\ (reasoning)\end{tabular}} & \multicolumn{1}{c}{\begin{tabular}[c]{@{}c@{}}Avg. CA\\ (trend)\end{tabular}} \\ \midrule
			\multirow{7}{*}{TempoGPT GPT-2 Linear}              & 1         & 64.3                               & 69.6                               & 64.5                                & 76.6                                & 61.7                                & 67.4                        & 67.6                                    & 67.0                                \\
	& 2         & 85.3                               & 69.6                               & 59.2                                & 77.9                                & 67.9                                & 72.0                        & 68.3                                    & 77.5                                \\
	& 3         & 95.8                               & 86.4                               & 68.4                                & 77.9                                & 65.4                                & 78.8                        & 70.6                                    & 91.1                                \\
	& 4         & 95.8                               & 89.6                               & 71.1                                & 77.9                                & 69.1                                & 80.7                        & 72.7                                    & 92.7                                \\
	& 5         & 95.8                               & 96.0                               & 71.1                                & 80.5                                & 67.9                                & 82.3                        & 73.2                                    & 95.9                                \\
	& 6         & 97.9                               & 91.2                               & 64.5                                & 68.8                                & 69.1                                & 78.3                        & 67.5                                    & 94.6                                \\
	& 7         & 97.2                               & 92.0                               & 69.7                                & 76.6                                & 72.8                                & 81.7                        & 73.1                                    & 94.6                                \\ \midrule

		\multirow{7}{*}{w/o quantization}                 & 1         & 40.6                               & 48.8                               & 51.3                                & 33.8                                & 17.3                                & 38.3                        & 34.1                                    & 44.7                                \\
		& 2         & 74.1                               & 78.4                               & 46.1                                & 31.2                                & 16.0                                & 49.2                        & 31.1                                    & 76.3                                \\
		& 3         & 90.2                               & 93.6                               & 51.3                                & 20.8                                & 16.0                                & 54.4                        & 29.4                                    & 91.9                                \\
		& 4         & 93.0                               & 85.6                               & 55.3                                & 54.5                                & 23.5                                & 62.4                        & 44.4                                    & 89.3                                \\
		& 5         & 98.6                               & 92.0                               & 71.1                                & 37.7                                & 44.4                                & 68.8                        & 51.1                                    & 95.3                                \\
		& 6         & 97.2                               & 91.2                               & 69.7                                & 45.5                                & 38.3                                & 68.4                        & 51.2                                    & 94.2                                \\
		& 7         & 97.2                               & 94.4                               & 71.1                                & 67.5                                & 40.7                                & 74.2                        & 59.8                                    & 95.8                                \\ \midrule

		\multirow{7}{*}{w/o pretrain} & 1         & 24.5                               & 24.8                               & 51.3                                & 40.3                                & 17.3                                & 31.6                        & 36.3                                    & 24.6                                \\
		& 2         & 32.9                               & 56.0                               & 32.9                                & 39.0                                & 18.5                                & 35.8                        & 30.1                                    & 44.4                                \\
		& 3         & 63.6                               & 68.0                               & 61.8                                & 46.8                                & 43.2                                & 56.7                        & 50.6                                    & 65.8                                \\
		& 4         & 78.3                               & 81.6                               & 71.1                                & 75.3                                & 63.0                                & 73.9                        & 69.8                                    & 80.0                                \\
		& 5         & 82.5                               & 88.0                               & 71.1                                & 72.7                                & 64.2                                & 75.7                        & 69.3                                    & 85.3                                \\
		& 6         & 85.3                               & 87.2                               & 71.1                                & 79.2                                & 60.5                                & 76.7                        & 70.3                                    & 86.3                                \\
		& 7         & 90.9                               & 90.4                               & 71.1                                & 79.2                                & 63.0                                & 78.9                        & 71.1                                    & 90.7                                \\ \bottomrule
	\end{tabular}}
\end{table*}

\end{document}